\documentclass[10pt,twocolumn,letterpaper]{article}

\usepackage{cvpr}              
\usepackage{amsmath}
\usepackage{bm}
\usepackage[table,dvipsnames]{xcolor}
\usepackage{multirow}
\usepackage{collcell}
\usepackage{subcaption}
\usepackage{array}
\usepackage{arydshln}
\usepackage{algorithm}
\usepackage{algpseudocode}
\usepackage{algorithmicx}
\usepackage{listings}
\usepackage{enumitem}
\usepackage{etoolbox}
\usepackage[accsupp]{axessibility} 
\makeatletter
\AfterEndEnvironment{algorithm}{\let\@algcomment\relax}
\AtEndEnvironment{algorithm}{\kern2pt\hrule\relax\vskip3pt\@algcomment}
\let\@algcomment\relax
\newcommand\algcomment[1]{\def\@algcomment{\footnotesize#1}}
\renewcommand\fs@ruled{\def\@fs@cfont{\bfseries}\let\@fs@capt\floatc@ruled
  \def\@fs@pre{\hrule height.8pt depth0pt \kern2pt}%
  \def\@fs@post{}%
  \def\@fs@mid{\kern2pt\hrule\kern2pt}%
  \let\@fs@iftopcapt\iftrue}
\makeatother

\makeatletter
\newcommand{\thickhline}{%
    \noalign {\ifnum 0=`}\fi \hrule height 1pt
    \futurelet \reserved@a \@xhline
}

\newcommand{\spthickhline}{%
    \noalign {\ifnum 0=`}\fi \hrule height 1.5pt
    \futurelet \reserved@a \@xhline
}
\newcommand{\sspthickhline}{%
    \noalign {\ifnum 0=`}\fi \hrule height 1.6pt
    \futurelet \reserved@a \@xhline
}
\definecolor{lightred}{RGB}{255,204,203}

\definecolor{cvprblue}{rgb}{0.21,0.49,0.74}
\usepackage[pagebackref,breaklinks,colorlinks,citecolor=cvprblue]{hyperref}

\def\paperID{5426} 
\def\confName{CVPR}
\def\confYear{2024}

\title{Poly Kernel Inception Network for Remote Sensing Detection}

\author{Xinhao Cai$^{1}$\thanks{Equal contribution.}~,~~\hspace{1pt}Qiuxia Lai$^{2}$\footnotemark[1],~~\hspace{1pt}Yuwei Wang$^{1}$\footnotemark[1],~~\hspace{1pt}Wenguan Wang$^{3}$\thanks{Corresponding author.},~~\hspace{1pt}Zeren Sun$^{1}$,~~~\hspace{1pt}Yazhou Yao$^{1}$\footnotemark[2]   \\
	\small {$^1$} Nanjing University of Science and Technology
	{$^2$}  Communication University of China
	{$^3$}  Zhejiang University \\
	\small \url{https://github.com/NUST-Machine-Intelligence-Laboratory/PKINet}
}

\begin{document}
\maketitle
\begin{abstract}

Object detection in remote sensing images (RSIs) often suffers from several increasing challenges, including the large variation in object scales and the diverse-ranging context. 
Prior methods tried to address these challenges by expanding the spatial receptive field of the backbone, either through large-kernel convolution or dilated convolution. However, the former typically introduces considerable background noise, while the latter risks generating overly sparse feature representations. 
In this paper, we introduce the Poly Kernel Inception Network (PKINet) to handle the above challenges.
PKINet employs multi-scale convolution kernels without dilation to extract object features of varying scales and capture local context. In 
addition, a Context Anchor Attention (CAA) module is introduced in parallel to capture long-range contextual information. 
These two components work jointly to advance the performance of PKINet on four challenging remote sensing detection benchmarks, namely DOTA-v1.0, DOTA-v1.5, HRSC2016, and DIOR-R.

\end{abstract}    

\vspace{-8pt}
\section{Introduction}
\label{sec:intro}

Object detection in remote sensing images (RSIs) has gained substantial attention in recent years~\cite{xia2018dota, ding2021object, sun2022fair1m}. This task is dedicated to discerning the presence of specific objects within RSIs and subsequently ascertaining their categories and precise locations. 
In contrast to generic object detection that typically produces horizontal bounding boxes, remote sensing object
detection aims to generate bounding boxes that align accurately with the orientation of the objects. 
Consequently, numerous prior efforts have been dedicated to developing various oriented bounding box (OBB) detectors~\cite{xu2020gliding, li2022oriented, xie2021oriented, ding2019learning, yang2021r3det, han2021align} and improving the angle prediction accuracy for OBBs~\cite{yang2021dense, yang2020arbitrary,yang2021rethinking, yang2021learning, yang2022kfiou}. Nevertheless, the unique characteristics of RSIs remain relatively under-explored when it comes to improving the feature extraction for object detection.

\begin{figure}[t!]
\vspace{-5pt}
\begin{center}
    \includegraphics[width=1\linewidth]{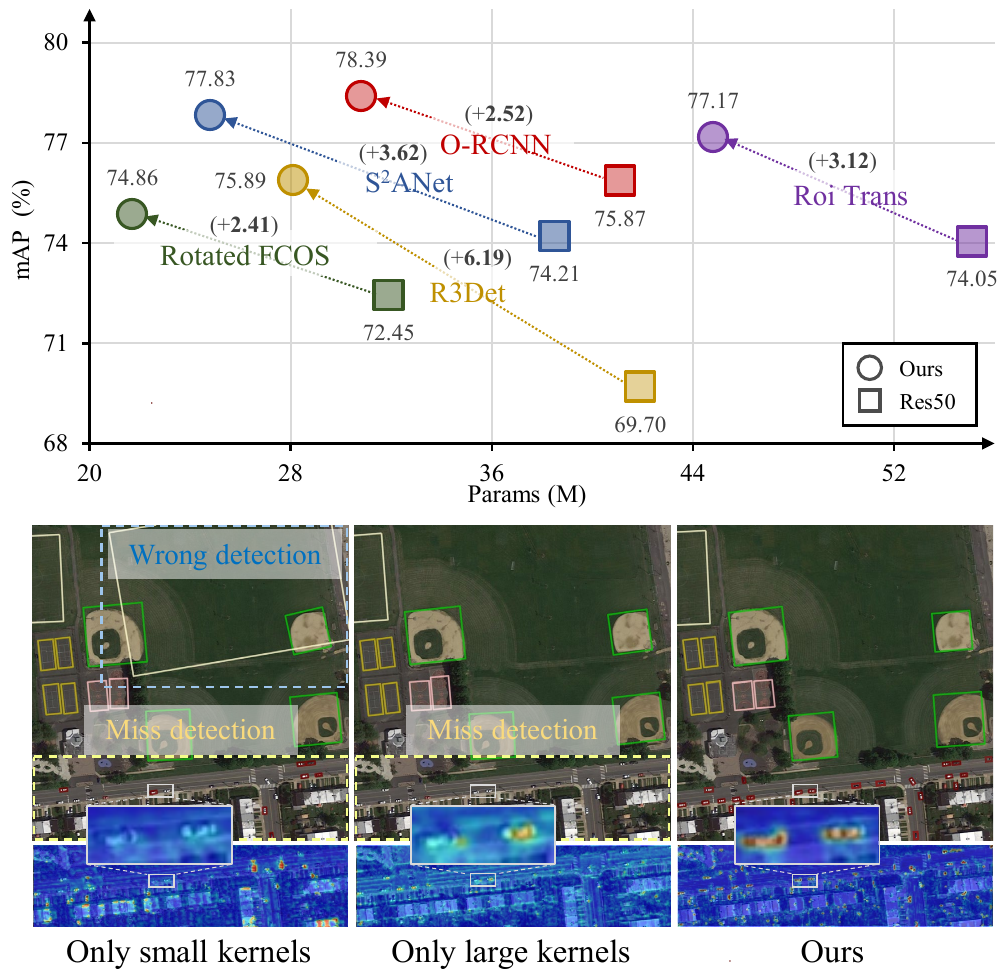}
        \put(-158,168){\scriptsize\cite{tian2019fcos}}
        \put(-116,158.5){\scriptsize\cite{yang2021r3det}}
        \put(-128.5,185){\scriptsize\cite{han2021align}}
        \put(-105.5,194){\scriptsize\cite{xie2021oriented}}
        \put(-21.5,181.5){\scriptsize\cite{ding2019learning}}
    \vspace{-0.3cm}
    \caption{\textbf{Top}: Our approach yields solid performance gains over various remote sensing detectors \cite{tian2019fcos, yang2021r3det, han2021align, xie2021oriented, ding2019learning} with fewer parameters on DOTA-v1.0 \cite{xia2018dota}. \textbf{Bottom}: Networks with small kernels miss long-range context in large object detection, whereas those with large kernels introduce noise for small objects. Our multi-scale convolution, however,  handles scale variations well.}
    \label{fig:figure1}
\end{center}
\vspace{-30pt}
\end{figure}

RSIs, including aerial and satellite images, are typically acquired from a bird's-eye perspective, offering high-resolution views of the Earth's surface. Consequently, objects depicted in RSIs exhibit a wide range of scales, spanning from expansive ones like soccer fields to relatively diminutive entities such as vehicles. Furthermore, the accurate recognition of these objects relies not solely on their appearances, but also on their contextual information, \textit{i.e.}, the surrounding environment in which they are situated. 
To address the large variation in the object scales, some methods employ explicit data augmentation techniques~\cite{zhao2019multi, shamsolmoali2021rotation, chen2020stitcher} to improve the robustness of the features against scale variations. Some resort to multi-scale feature integration~\cite{zhang2019hierarchical, lin2018squeeze} or pyramidal feature hierarchy~\cite{liang2019small, 9382268} to extract features rich in scale information. Nevertheless, a limitation of these methods is that the receptive fields for objects of varying scales remain identical, thereby failing to provide sufficient contextual information for larger objects.

Recently, LSKNet~\cite{Li_2023_ICCV} proposes to selectively enlarge the spatial receptive field for larger objects to capture more scene context information. This is achieved by incorporating large-kernel convolutions~\cite{liu2022convnet, guo2023visual, ding2022scaling, liu2022more} and dilated convolutions into the backbone network. However, it is noteworthy that the use of large-kernel convolutions may introduce a significant amount of background noise, which could be detrimental to the accurate detection of small objects. On the other hand, dilated convolutions, though effective at enlarging the receptive field, might inadvertently overlook fine-grained details within that field, potentially resulting in overly sparse feature representations. 

To address the challenges posed by the large variation in object scales and the diverse-ranging context within RSIs, in this paper, we present a powerful and lightweight feature extraction backbone network named Poly Kernel Inception Network (PKINet) for remote sensing object detection.  Unlike previous methods that rely on large-kernel or dilated convolutions to expand the receptive field, PKINet arranges multiple depth-wise convolution kernels of different sizes without dilation in parallel, and extracts dense texture features across varying receptive fields. These texture features are adaptively fused along the channel dimension, enabling the collection of local contextual information. To further encompass long-range contextual information, we introduce a Context Anchor Attention (CAA) mechanism, which leverages global average pooling and 1D strip convolutions to capture the relationships between distant pixels and enhances the features within the central region. The two components work jointly to facilitate the extraction of adaptive features with both local and global contextual information, thereby improving the performance of remote sensing object detection.


To the best of our knowledge, PKINet represents the pioneering effort in exploring the application of inception-style convolutions and global context attention in remote sensing object detection, aiming to effectively tackle the challenges posed by the considerable variations in object scale and contextual diversity. 
Extensive experiments on widely used remote sensing benchmarks DOTA-v1.0 \cite{xia2018dota}, DOTA-v1.5 \cite{xia2018dota}, HRSC2016 \cite{liu2017high}, and DIOR-R \cite{cheng2022anchor} demonstrate the effectiveness of our method. 
In addition to its exceptional feature extraction capabilities, our model is lightweight compared with previous methods thanks to the strategic use of depth-wise and 1D convolutions.

\section{Related Work}
The challenges faced by remote sensing object detection primarily stem from objects with arbitrary orientations and substantial scale variations~\cite{xia2018dota, ding2021object, sun2022fair1m, liu2016ship, cheng2022anchor, yao2023automated}. The majority of previous methods have focused on oriented bounding box (OBB) detection. Nonetheless, an emerging trend is to design effective feature extraction backbones tailored to the characteristics of remote sensing images (RSIs).

\noindent\textbf{OBB for Remote Sensing Object Detection.}
To address the challenge of arbitrary orientations of the objects in RSIs, one research direction focuses on \textit{developing specialized OBB detectors}. This includes introducing feature refinement techniques into the detector neck~\cite{yang2019scrdet,yang2021r3det}, extracting the rotated region of interest (RoI)~\cite{ding2019learning,xie2021oriented}, designing specific detection heads for the OBBs~\cite{hou2022shape,han2021redet, ming2021dynamic},~\textit{.etc}.  
Though improving over general horizontal bounding box (HBB) detectors, these methods often suffer from issues like boundary discontinuity due to their relatively inflexible object representations obtained by augmenting horizontal object representations with additional angle parameters.
To mitigate the aforementioned issues, another line of research has been dedicated to \textit{developing new object representations} for detecting OBBs \cite{xu2020gliding, li2022oriented, wang2019mask, yi2021oriented, fu2020point, yang2021dense}. For example, Xu~\textit{et al}.~\cite{xu2020gliding} propose to describe a multi-oriented object by adding four gliding offset variables to classical HBB representation.
Li~\textit{et al}.~\cite{li2022oriented} characterize oriented objects using a set of points to achieve more accurate orientation estimation.
Some others~\cite{hou2023g, yang2021rethinking, yang2021learning, cheng2022dual} utilize Gaussian distributions to model the OBBs for object detection and design new loss functions \cite{qian2021learning} to guide the learning process.

Although these methods are promising in addressing challenges related to arbitrary orientations, they typically rely on standard backbones for feature extraction, which often overlook the unique characteristics of RSIs that are essential for object detection, \textit{e.g.}, the large object scale variations and the diverse contextual information. 
In contrast, we propose a feature extraction backbone to deal with the challenges posed by the large object scale variations.

\begin{figure*}[t!]
\vspace{-10pt}
    \centering    \includegraphics[width=1\linewidth]{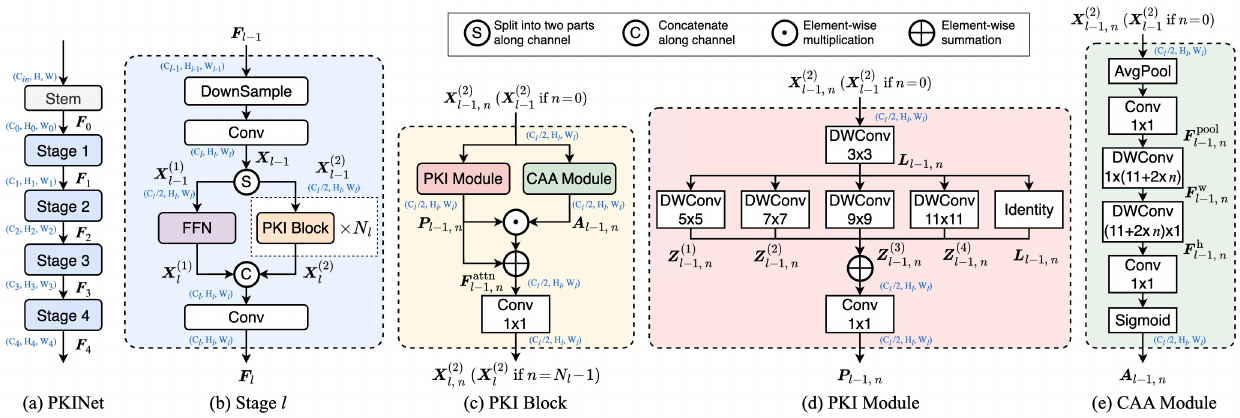}
    \vspace{-22pt}
    \caption{\textbf{PKINet overview.} \textbf{(a) PKINet} consists of four stages, where the spatial resolution of the $l$-th stage output is $(C_l\!\times\!H_l\!\times\!W_l)$. Each \textbf{(b) Stage} (\S\ref{sec:pki_stage}) implies a cross-stage partial (CSP) structure, where the input is split in half along the channel dimension and fed to a Feed-Forward Network (FFN) and a sequence of $N_l$ PKI Blocks, respectively. Each \textbf{(c) PKINet Block} contains a \textbf{(d) PKI Module} (\S\ref{sec:pki_module}) and a \textbf{(e) CAA Module} (\S\ref{sec:caa}). 
    Here, $n\!=\!0,\dots,N_l\!-\!1$ means that the PKI/CAA Module is in the $n$-th PKI Block of the $l$-th stage.
    }
    \vspace{-10pt}
    \label{fig:framework}
\end{figure*}

\noindent\textbf{Feature Extraction for Remote Sensing Object Detection.}
To better handle the unique challenges such as large object scale variations in RSIs, certain methods emphasize the \textit{extraction of multi-scale features} through approaches like data augmentation~\cite{zhao2019multi, shamsolmoali2021rotation, chen2020stitcher}, multi-scale feature integration~\cite{zhang2019hierarchical,zheng2020hynet,9382268,liu2016ssd}, feature pyramid network (FPN) enhancement~\cite{lin2017feature,hou2022refined,zhang2021laplacian, guo2020rotational}, or multi-scale anchor generation~\cite{guo2018geospatial, hou2021self, qiu2019a2rmnet}. 
Recently, there has been a noteworthy development in the \textit{design of feature extraction backbones} specifically for remote sensing object detection. Some~\cite{han2021redet,pu2023adaptive} focus on extracting features suitable for objects of varying orientations with equivalent receptive fields. Some~\cite{Li_2023_ICCV} enlarge the spatial receptive field for larger objects using large kernels~\cite{liu2022convnet,ding2022scaling,liu2022more}, which inevitably introduces background noise for smaller objects. Some~\cite{guo2022segnext,dai2023multi,zhang2022multiscale} adopt multi-scale convolution kernels in order to address challenges across various fields, yet research in remote sensing detection remains scarce.

Similar to~\cite{Li_2023_ICCV}, we propose a new feature extraction backbone PKINet to address the challenges posed by the large variation in object scales and diverse context in RSIs. There are \textbf{two key differences} between the two methods. Firstly, instead of relying on large-kernel or dilated convolutions to expand the receptive field, PKINet utilizes inception-style depth-wise convolution without dilation to extract multi-scale texture features across varying receptive fields. Secondly, our method incorporates a Context Anchor Attention (CAA) mechanism to capture the long-range contextual information. The two components collaborate to facilitate the extraction of adaptive features with both local and global contextual information, thereby improving the performance of remote sensing object detection.

\vspace{-3pt}
\section{Methodology}
\vspace{-2pt}
As shown in Fig.$_{\!}$~\ref{fig:framework}(a), our PKINet is a feature extraction backbone similar to VGG~\cite{simonyan2014very} and ResNet~\cite{he2016deep}, which consists of four stages.
Each stage (\S\ref{sec:pki_stage}) implies a cross-stage partial (CSP) structure~\cite{wang2020cspnet}, where the stage input is split and fed into two paths. 
One path is a simple Feed-Forward Network (FFN). 
The other path consists of a sequence of PKI Blocks, and each PKI Block contains a PKI Module (\S\ref{sec:pki_module}) and a CAA Module (\S\ref{sec:caa}).
The outputs of the two paths are concatenated to yield the output of the stage.
PKINet can be incorporated with various oriented object detectors such as Oriented RCNN~\cite{xie2021oriented} to produce the final object detection results for RSIs.

\vspace{-1pt}
\subsection{PKI Stage}
\label{sec:pki_stage}
\vspace{-1pt}
There are four stages arranged sequentially in PKINet. The input and output of stage $l$ are $\bm{F}_{l-1\!}\!\in\!\mathbb{R}^{C_{l-1}\!\times\!H_{l-1}\!\times\!W_{l-1}}$ and $\bm{F}_{l}\!\in\!\mathbb{R}^{C_l\!\times\!H_l\!\times\!W_l}$, respectively. 
The structure of stage $l$ is shown in Fig.~\ref{fig:framework}(b), which implies a cross-stage partial (CSP) structure~\cite{wang2020cspnet}. Specifically, the stage input $\bm{F}_{l-1}$ after initial processing is split in half along the channel dimension and fed into two paths: 
\begin{equation}\small
    \begin{split}
    \bm{X}_{l-1} = \texttt{Conv}_{3\!\times\!3}(\texttt{DS}(\bm{F}_{l-1})) &\in\!\mathbb{R}^{C_l\!\times\!H_l\!\times\!W_l}, \\
    \bm{X}^{(1)}_{l-1} = \bm{X}_{l-1}[:\frac{1}{2} C_{l},\dots],~\bm{X}^{(2)}_{l-1}& = \bm{X}_{l-1}[\frac{1}{2} C_{l}:,\dots], 
    \end{split}
        \vspace{-4pt}
\end{equation}
where \texttt{DS} denotes the downsampling operation. 
One path is a simple Feed-Forward Network (FFN), which takes in $\bm{X}^{(1)\!}_{l-1}\!\in\!\mathbb{R}^{\frac{1}{2} C_l\!\times\!H_l\!\times\!W_l\!}$ and then output $\bm{X}^{(1)\!}_{l}\!\in\!\mathbb{R}^{\frac{1}{2} C_l\!\times\!H_l\!\times\!W_l}$.
The other path consists of a sequence of $N_l$ PKI Blocks, which processes $\bm{X}^{(2)\!}_{l-1}\!\in\!\mathbb{R}^{\frac{1}{2} C_l\!\times\!H_l\!\times\!W_l}$ and yields $\bm{X}^{(2)\!}_{l}\!\in\!\mathbb{R}^{\frac{1}{2} C_l\!\times\!H_l\!\times\!W_l}$. 
As shown in Fig.$_{\!}$~\ref{fig:framework} (c), PKI Block contains a PKI Module and a CAA Module, which will be detailed in~\S\ref{sec:pki_module} and~\S\ref{sec:caa}, respectively.
The final output of stage $l$ is:
\vspace{-5pt}
\begin{equation}\small
    \bm{F}_{l}\!=\!\texttt{Conv}_{1\!\times\!1}(\texttt{Concat}(\bm{X}^{(1)}_{l}, \bm{X}^{(2)}_{l}))\in\!\mathbb{R}^{C_l\!\times\!H_l\!\times\!W_l},
\end{equation}
where \texttt{Concat} refers to the concatenation operation.

\subsection{PKI Module}
\label{sec:pki_module}
A PKINet Block consists of a PKI Module and a CAA Module. In this section, we look into the details of PKI Module. We will introduce CAA Module in~\S\ref{sec:caa}.

As discussed in~\S\ref{sec:intro}, different from general object detection, remote sensing object detection aims to locate and recognize objects of varying sizes within a single image. To address the challenges related to large variations in object scales, we introduce \textit{PKI Module} to capture multi-scale texture features. As showcased in Fig.~\ref{fig:framework}(d), PKI Module is an inception-style module~\cite{yu2023inceptionnext, szegedy2016rethinking} that comprises a small-kernel convolution to grasp local information, followed by a set of parallel depth-wise convolutions to capture contextual information across multiple scales. 
Formally, PKI Module within the $n$-th PKI Block of the $l$-th stage can be represented mathematically as follows:
\vspace{-3pt}
\begin{equation}\small
    \begin{split}
    &\bm{L}_{l-1,~n} = \texttt{Conv}_{k_s\!\times k_s}(\bm{X}^{(2)}_{l-1,~n}),~n\!=\!0,\dots,N_l\!-\!1, \\
    &\bm{Z}_{l-1,~n}^{(m)} = \texttt{DWConv}_{k^{(m)}\!\times k^{(m)}} (\bm{L}_{l-1,~n}), ~m\!=\!1,\dots,4. 
    \end{split}
    \vspace{-3pt}
\end{equation}
Here, $\bm{L}_{l-1,~n\!}\!\in\!\mathbb{R}^{\frac{1}{2} C_l\!\times\!H_l\!\times\!W_l\!}$ is the local feature extracted by the $k_s\!\times\! k_s$ convolution, and $\bm{Z}_{l-1,~n}^{(m)\!}\!\in\!\mathbb{R}^{\frac{1}{2} C_l\!\times\!H_l\!\times\!W_l\!}$ is the context feature extracted by the $m$-th $k^{(m)}\!\times\!k^{(m)}$ depth-wise convolution (\texttt{DWConv}). 
In our experiment, we set $k_s\!=\!3$ and $k^{(m)}\!=\!(m\!+\!1)\!\times\!2\!+\!1$. 
For $n\!=\!0$, we have $\bm{X}^{(2)}_{l-1,~n}\!=\!\bm{X}^{(2)}_{l-1}$.
Note that our PKI Module does not use dilated convolution, thereby preventing the extraction of overly sparse feature representations.

Then, the local and contextual features are fused by a convolution of size $1\!\times\!1$, characterizing the interrelations among various channels: 
\vspace{-4pt}
\begin{equation}\small
\label{eq:pki_module_output}
    \bm{P}_{l-1,~n} = \texttt{Conv}_{1\!\times\!1}\big(\bm{L}_{l-1,~n} + \sum\nolimits_{m=1}^{4} \bm{Z}_{l-1,~n}^{(m)}\big),
    \vspace{-3pt}
\end{equation}
where $\bm{P}_{l-1,~n}\!\in\!\mathbb{R}^{\frac{1}{2} C_l\!\times\!H_l\!\times\!W_l\!}$ represents the output feature.
The ${1\!\times\!1}$ convolution serves as a channel fusion mechanism to integrate features with varying receptive field sizes. In this way, our PKI Module could capture a broad spectrum of contextual information without compromising the integrity of local texture features.

\subsection{Context Anchor Attention (CAA)}
\label{sec:caa}

As discussed above, the inception-style PKI Module in the PKI Block focuses on extracting multi-scale local contextual information. 
To capture long-range contextual information, inspired by~\cite{tang2022ghostnetv2, Li_2023_ICCV}, we further integrated a Context Anchor Attention (CAA) module into the PKI Block. 
CAA aims to grasp contextual interdependencies among distant pixels while augmenting the central features concurrently. An illustration of CAA is presented in Fig.~\ref{fig:framework}(e).
Take the CAA Module in the $n$-th PKI Block of the $l$-th stage as an example, we adopt average pooling followed by a $1\!\times\!1$ convolution to obtain the local region feature:
\vspace{-3pt}
\begin{equation}\small
    \bm{F}_{l-1,~n}^\text{pool} = \texttt{Conv}_{1\!\times\!1}(\mathcal{P}_{avg}(\bm{X}^{(2)}_{l-1,~n})),~n\!=\!0,\dots,N_l\!-\!1,
    \vspace{-2pt}
\end{equation}
where $\mathcal{P}_{avg}$ represents average pooling operation. For $n\!=\!0$, we have $\bm{X}^{(2)}_{l-1,~n\!}\!=\!\bm{X}^{(2)}_{l-1}$. Then, we apply two depth-wise strip convolutions as an approximation to a standard large-kernel depth-wise convolution:
    \vspace{-3pt}
\begin{equation}\small
    \begin{split}
    \bm{F}_{l-1,~n}^\text{w} &= \texttt{DWConv}_{1\!\times k_{b}} (\bm{F}_{l-1,~n}^\text{pool}), \\
    \bm{F}_{l-1,~n}^\text{h} &= \texttt{DWConv}_{k_{b} \times\!1} (\bm{F}_{l-1,~n}^\text{w}).
    \end{split}
        \vspace{-2pt}
\end{equation}
We opt for depth-wise strip convolutions based on two primary considerations. First, strip convolution is lightweight. 
Compared to a conventional $k_{b}\!\times\!k_{b}$ 2D depth-wise convolution, we can achieve a similar effect with a couple of 1D depth-wise kernels with a parameter reduction of $k_{b}/2$. Second, strip convolution can facilitate the identification and extraction of features for objects with slender shapes, such as bridges. 
To increase the receptive field of CAA Module as the PKI Block it belongs to goes deeper, we set $k_{b}\!=\!11\!+\!2\!\times\!l$, \textit{i.e.}, we calculate the kernel size $k_{b}$ as the function of the PKI Block depth $n$. Such a design enhances the ability of PKINet to establish the relationship between long-range pixels, and would not significantly increase the computational cost thanks to the strip depth-wise design.

Finally, our CAA Module produces an attention weight

\noindent $\bm{A}_{l-1,~n\!}\!\in\!\mathbb{R}^{\frac{1}{2} C_l\!\times\!H_l\!\times\!W_l}$, which is further used to enhance the output of PKI Module (\textit{cf.} Eq.~(\ref{eq:pki_module_output})):
 \vspace{-2pt}
\begin{equation}\small
    \begin{split}
    \bm{A}_{l-1,~n} &= \texttt{Sigmoid}(\texttt{Conv}_{1\!\times\!1}(\bm{F}_{l-1,~n}^\text{h})), \\
    \bm{F}_{l-1,~n}^\text{attn} &= (\bm{A}_{l-1,~n} \odot \bm{P}_{l-1,~n}) \oplus \bm{P}_{l-1,~n}.
    \end{split}
     \vspace{-2pt}
\end{equation}
Here, $\texttt{Sigmoid}$ function ensures that the attention map $\bm{A}_{l-1,~n}$ is in range $(0, 1)$, $\odot$ denotes the element-wise multiplication, $\oplus$ denotes the element-wise summation, and $\bm{F}_{l-1,~n\!}^\text{attn}\!\in\!\mathbb{R}^{\frac{1}{2} C_l\!\times\!H_l\!\times\!W_l\!}$ is the enhanced feature. 
The output of the $n$-th PKI Block in the $l$-th stage is obtained by:
 \vspace{-2pt}
\begin{equation}\small
    \bm{X}^{(2)}_{l,~n} = \texttt{Conv}_{1\!\times\!1}(\bm{F}_{l-1,~n}^\text{attn}).
     \vspace{-2pt}
\end{equation}
For $n\!=\!N_l\!-\!1$, we have $\bm{X}^{(2)}_{l}\!=\!\bm{X}^{(2)}_{l,~n}$, \textit{i.e.}, we denote the output of the last PKI Block as $\bm{X}^{(2)}_{l}$.

\begin{table}[t!]
\resizebox{\linewidth}{!}{
			\setlength\tabcolsep{8pt}
			\renewcommand\arraystretch{1.05}
\hspace{-1.0em}
\begin{tabular}{cccc||cc}
\spthickhline
\multicolumn{1}{c|}{\multirow{2}{*}{}} & \multicolumn{1}{c|}{\multirow{2}{*}{$\frac{H_l}{H}\!\times\!\frac{W_l}{W}$}}  & \multicolumn{2}{c||}{\multirow{2}{*}{Layer Specification}}                                                    & \multicolumn{2}{c}{PKINet}         \\ \cline{5-6} 
\multicolumn{1}{c|}{}                       & \multicolumn{1}{c|}{}                             & \multicolumn{2}{c||}{}                                                                                        & \multicolumn{1}{c|}{T}     & S     \\ \hline \hline
\multicolumn{1}{c|}{\multirow{1}{*}{Stem}} & \multicolumn{1}{c|}{\multirow{1}{*}{$\frac{1}{2} \times \frac{1}{2}$}} & \multicolumn{1}{c|}{\multirow{1}{*}{\begin{tabular}[c]{@{}c@{}}Down-samp.\end{tabular}}} & Kernel Size & \multicolumn{2}{c}{$3\times3$, stride 2} \\ \hline
\multicolumn{1}{c|}{\multirow{4}{*}{Stage 1}}     & \multicolumn{1}{c|}{\multirow{4}{*}{$\frac{1}{4} \times \frac{1}{4}$}}   & \multicolumn{1}{c|}{\multirow{2}{*}{\begin{tabular}[c]{@{}c@{}}Down-\\ sampling\end{tabular}}} & Kernel Size & \multicolumn{2}{c}{$3\times3$, stride 2}  \\ \cline{4-6} 
\multicolumn{1}{c|}{}                       & \multicolumn{1}{c|}{}                             & \multicolumn{1}{c|}{}                                                                          & Embed. Dim  & \multicolumn{1}{c|}{32}    & 64    \\ \cline{3-6} 
\multicolumn{1}{c|}{}                       & \multicolumn{1}{c|}{}                             & \multicolumn{1}{c|}{\multirow{2}{*}{\begin{tabular}[c]{@{}c@{}}PKI\\ Block\end{tabular}}}   & Kernel Size & \multicolumn{2}{c}{$3\times3$ to $11\times11$}  \\ \cline{4-6} 
\multicolumn{1}{c|}{}                       & \multicolumn{1}{c|}{}                             & \multicolumn{1}{c|}{}                                                    & \#Block ($N_1$)       & \multicolumn{2}{c}{4}              \\ \hline
\multicolumn{1}{c|}{\multirow{4}{*}{Stage 2}}     & \multicolumn{1}{c|}{\multirow{4}{*}{$\frac{1}{8} \times \frac{1}{8}$}}   & \multicolumn{1}{c|}{\multirow{2}{*}{\begin{tabular}[c]{@{}c@{}}Down-\\ sampling\end{tabular}}} & Kernel Size & \multicolumn{2}{c}{$3\times3$, stride 2}  \\ \cline{4-6} 
\multicolumn{1}{c|}{}                       & \multicolumn{1}{c|}{}                             & \multicolumn{1}{c|}{}                                                                          & Embed. Dim  & \multicolumn{1}{c|}{64}    & 128   \\ \cline{3-6} 
\multicolumn{1}{c|}{}                       & \multicolumn{1}{c|}{}                             & \multicolumn{1}{c|}{\multirow{2}{*}{\begin{tabular}[c]{@{}c@{}}PKI\\ Block\end{tabular}}}   & Kernel Size & \multicolumn{2}{c}{$3\times3$ to $11\times11$}  \\ \cline{4-6} 
\multicolumn{1}{c|}{}                       & \multicolumn{1}{c|}{}                             & \multicolumn{1}{c|}{}                        & \#Block ($N_2$)       & \multicolumn{1}{c|}{14}    & 12    \\ \hline
\multicolumn{1}{c|}{\multirow{4}{*}{Stage 3}}     & \multicolumn{1}{c|}{\multirow{4}{*}{$\frac{1}{16} \times \frac{1}{16}$}} & \multicolumn{1}{c|}{\multirow{2}{*}{\begin{tabular}[c]{@{}c@{}}Down-\\ sampling\end{tabular}}} & Kernel Size & \multicolumn{2}{c}{$3\times3$, stride 2}  \\ \cline{4-6} 
\multicolumn{1}{c|}{}                       & \multicolumn{1}{c|}{}                             & \multicolumn{1}{c|}{}                                                                          & Embed. Dim  & \multicolumn{1}{c|}{128}   & 256   \\ \cline{3-6} 
\multicolumn{1}{c|}{}                       & \multicolumn{1}{c|}{}                             & \multicolumn{1}{c|}{\multirow{2}{*}{\begin{tabular}[c]{@{}c@{}}PKI\\ Block\end{tabular}}}   & Kernel Size & \multicolumn{2}{c}{$3\times3$ to $11\times11$}  \\ \cline{4-6} 
\multicolumn{1}{c|}{}                       & \multicolumn{1}{c|}{}                             & \multicolumn{1}{c|}{}                       & \#Block ($N_3$)       & \multicolumn{1}{c|}{22}    & 20    \\ \hline
\multicolumn{1}{c|}{\multirow{4}{*}{Stage 4}}     & \multicolumn{1}{c|}{\multirow{4}{*}{$\frac{1}{32} \times \frac{1}{32}$}} & \multicolumn{1}{c|}{\multirow{2}{*}{\begin{tabular}[c]{@{}c@{}}Down-\\ sampling\end{tabular}}} & Kernel Size & \multicolumn{2}{c}{$3\times3$, stride 2}  \\ \cline{4-6} 
\multicolumn{1}{c|}{}                       & \multicolumn{1}{c|}{}                             & \multicolumn{1}{c|}{}                                                                          & Embed. Dim  & \multicolumn{1}{c|}{256}   & 512   \\ \cline{3-6} 
\multicolumn{1}{c|}{}                       & \multicolumn{1}{c|}{}                             & \multicolumn{1}{c|}{\multirow{2}{*}{\begin{tabular}[c]{@{}c@{}}PKI\\ Block\end{tabular}}}   & Kernel Size & \multicolumn{2}{c}{$3\times3$ to $11\times11$}  \\ \cline{4-6} 
\multicolumn{1}{c|}{}                       & \multicolumn{1}{c|}{}                             & \multicolumn{1}{c|}{}                     & \#Block ($N_4$)       & \multicolumn{2}{c}{4}              \\ \hline \hline
\multicolumn{4}{c||}{Parameters (M)}                                                                                                                                                                            & \multicolumn{1}{c|}{4.13}  & 13.69 \\ \hline
\multicolumn{4}{c||}{FLOPs (G)}                                                                                                                                                                                 & \multicolumn{1}{c|}{22.70} & 70.20 \\
\hline
\end{tabular}
}
\vspace{-8pt}
\caption{\textbf{Configurations of two variants of PKINet.} Here, ``T'' denotes ``Tiny'', and ``S'' denotes ``Small''. See \S\ref{sec:imp_details} for details.}
\label{tab:structure}
\vspace{-10pt}
\end{table}

\subsection{Implementation Details}
\label{sec:imp_details}
In this paper, we present two variants of the proposed backbone, namely \textbf{PKINet-T} and \textbf{PKINet-S}, where ``T'' stands for ``Tiny'', and ``S'' stands for ``Small''. 
The Stem structure consists of three $3\!\times\!3$ convolution layers with strides $(2,~1,~1)$, respectively. 
For both PKINet-T and PKINet-S, $H_l\!=\!H/2^{(l\!+\!1)}, W_l\!=\!W/2^{(l\!+\!1)}$ for $l\!=\!0,\dots,4$, and $H, W$ are the height and width of the input, respectively. 
For PKINet-T, $C_0\!=\!32, C_l\!=\!2^{l\!-\!1}\!\times\!C_0$ for $l\!=\!1,\dots,4$, and the number of PKI Blocks of the four stages are $(4, 14, 22, 4)$, respectively.
For PKINet-S, $C_0\!=\!64, C_l\!=\!2^{l\!-\!1}\!\times\!C_0$ for $l\!=\!1,\dots,4$, and the number of PKI Blocks of the four stages are $(4, 12, 20, 4)$, respectively. 
Please note that although PKINet-T comprises more PKI Blocks compared to PKINet-S, it contains significantly fewer parameters owing to the halved channel number in the intermediate features.
The detailed conﬁgurations of the two variants of PKINet are listed in Table~\ref{tab:structure}.

\begin{table*}[t]
 \vspace{-5pt}
\setlength{\tabcolsep}{2pt}
\renewcommand\arraystretch{1.15}
\scriptsize      
\centering
\resizebox{\textwidth}{!}{
\hspace{-1.0em}
\begin{tabular}{r|c||c|ccccccccccccccc||l} 
\thickhline
\rowcolor[rgb]{0.92,0.92,0.92} Method       & \textbf{Backbone}   & \#\textbf{P} $\downarrow$    & PL    & BD    & BR    & GTF   & SV    & LV    & SH    & TC    & BC    & ST    & SBF   & RA    & HA    & SP    & HC  &\textbf{mAP} $\uparrow$  \\ 
\hline
\hline
 \multicolumn{2}{l}{\textbf{\textit{DETR-based}}}\\
\hline
 AO\textsuperscript{2}-DETR~\cite{dai2022ao2}  & ResNet-50~\cite{he2016deep} & 40.8M & 87.99 & 79.46 & 45.74 & 66.64 & 78.90 & 73.90 & 73.30 & 90.40 & 80.55 & 85.89 & 55.19 & 63.62 & 51.83 & 70.15 & 60.04 & 70.91 \\ \hline
 O\textsuperscript{2}-DETR~\cite{ma2021oriented}  & ResNet-50~\cite{he2016deep} & - & 86.01 & 75.92 & 46.02 & 66.65 & 79.70 & 79.93 & 89.17 & 90.44 & 81.19 & 76.00 & 56.91 & 62.45 & 64.22 & 65.80 & 58.96 & 72.15 \\ \hline
 ARS-DETR~\cite{zeng2023ars}  & ResNet-50~\cite{he2016deep} & 41.6M  & 86.61 & 77.26 & 48.84 & 66.76 & 78.38 & 78.96 & 87.40 & 90.61 & 82.76 & 82.19 & 54.02 & 62.61 & 72.64 & 72.80 & 64.96 & 73.79 \\
\hline
\hline
\multicolumn{2}{l}{\textbf{\textit{One-stage}}}\\
\hline
SASM~\cite{hou2022shape}  & ResNet-50~\cite{he2016deep} & 36.6M & 86.42 & 78.97 & 52.47 & 69.84 & 77.30 & 75.99 & 86.72 & 90.89 & 82.63 & 85.66 & 60.13 & 68.25 & 73.98 & 72.22 & 62.37 & 74.92 \\ \hline
R3Det-GWD~\cite{yang2021rethinking} & ResNet-50~\cite{he2016deep} & 41.9M & 88.82 & 82.94 & 55.63 & 72.75 & 78.52 & 83.10 & 87.46 & 90.21 & 86.36 & 85.44 & 64.70 & 61.41 & 73.46 & 76.94 & 57.38 & 76.34 \\ \hline
R3Det-KLD~\cite{yang2021learning} & ResNet-50~\cite{he2016deep} & 41.9M & 88.90 & 84.17 & 55.80 & 69.35 & 78.72 & 84.08 & 87.00 & 89.75 & 84.32 & 85.73 & 64.74 & 61.80 & 76.62 & 78.49 & 70.89 & 77.36 \\ \hline
O-RepPoints~\cite{li2022oriented}  & ResNet-50~\cite{he2016deep} & 36.6M & 87.02 & 83.17 & 54.13 & 71.16 & 80.18 & 78.40 & 87.28 & 90.90 & 85.97 & 86.25 & 59.90 & 70.49 & 73.53 & 72.27 & 58.97 & 75.97 \\ \hline
Rotated & ResNet-50~\cite{he2016deep} & 31.9M & 88.52 & 77.54 & 47.06 & 63.78 & 80.42 & 80.50 & 87.34 & 90.39 & 77.83 & 84.13 & 55.45 & 65.84 & 66.02 & 72.77 & 49.17 & 72.45 \\
FCOS~\cite{tian2019fcos} & \textbf{PKINet-S} & \textbf{21.7}M & 88.56 & 82.89 & 47.96 & 58.20 & 81.09 & 83.09 & 88.23 & 90.88 & 84.57 & 85.81 & 57.98 & 66.26 & 75.12 & 80.93 & 51.39 & \textbf{74.86} \\
\hdashline
\multirow{3}{*}{R3Det~\cite{yang2021r3det}} & ResNet-50~\cite{he2016deep} & 41.9M & 89.00 & 75.60 & 46.64 & 67.09 & 76.18 & 73.40 & 79.02 & 90.88 & 78.62 & 84.88 & 59.00 & 61.16 & 63.65 & 62.39 & 37.94 & 69.70 \\
& ARC-R50~\cite{pu2023adaptive} & 65.2M & 89.49 & 78.04 & 46.36 & 68.89 & 77.45 & 72.87 & 82.76 & 90.90 & 83.07 & 84.89 & 58.72 & 68.61 & 64.75 & 68.39 & 49.67 & 72.32 \\
& \textbf{PKINet-S} & \textbf{28.1}M & 89.63 & 82.40 & 49.77 & 71.72 & 79.95 & 81.39 & 87.79 & 90.90 & 84.20 & 86.09 & 61.08 & 66.55 & 73.06 & 73.85 & 59.95 & \textbf{75.89} \\
\hdashline
\multirow{3}{*}{S$^2$ANet~\cite{han2021align}} & ResNet-50~\cite{he2016deep} & 38.5M & 89.11 & 82.84 & 48.37 & 71.11 & 78.11 & 78.39 & 87.25 & 90.83 & 84.90 & 85.64 & 60.36 & 62.60 & 65.26 & 69.13 & 57.94 & 74.12  \\
& ARC-R50~\cite{pu2023adaptive} & 71.8M & 89.28 & 78.77 & 53.00 & 72.44 & 79.81 & 77.84 & 86.81 & 90.88 & 84.27 & 86.20 & 60.74 & 68.97 & 66.35 & 71.25 & 65.77 & 75.49\\
& \textbf{PKINet-S} & \textbf{24.8}M & 89.67 & 84.16 & 51.94 & 71.89 & 80.81 & 83.47 & 88.29 & 90.80 & 87.01 & 86.94 & 65.02 & 69.53 & 75.83 & 80.20 & 61.85 & \textbf{77.83} \\
\hline
\hline
\multicolumn{2}{l}{\textbf{\textit{Two-stage}}}\\
\hline
SCRDet~\cite{yang2019scrdet} & ResNet-50~\cite{he2016deep} & 41.9M & 89.98 & 80.65 & 52.09 & 68.36 & 68.36 & 60.32 & 72.41 & 90.85 & 87.94 & 86.86 & 65.02 & 66.68 & 66.25 & 68.24 & 65.21 & 72.61 \\ \hline
G.V.~\cite{xu2020gliding}  & ResNet-50~\cite{he2016deep} & 41.1M & 89.64 & 85.00 & 52.26 & 77.34 & 73.01 & 73.14 & 86.82 & 90.74 & 79.02 & 86.81 & 59.55 & 70.91 & 72.94 & 70.86 & 57.32 & 75.02 \\ \hline
CenterMap~\cite{long2021creating} & ResNet-50~\cite{he2016deep} & 41.1M & 89.02 & 80.56 & 49.41 & 61.98 & 77.99 & 74.19 & 83.74 & 89.44 & 78.01 & 83.52 & 47.64 & 65.93 & 63.68 & 67.07 & 61.59 & 71.59 \\ \hline
ReDet~\cite{han2021redet}  & ResNet-50~\cite{he2016deep} & 31.6M & 88.79 & 82.64 & 53.97 & 74.00 & 78.13 & 84.06 & 88.04 & 90.89 & 87.78 & 85.75 & 61.76 & 60.39 & 75.96 & 68.07 & 63.59 & 76.25 \\ \hline
\multirow{2}{*}{Roi Trans \cite{ding2019learning}} & ResNet-50~\cite{he2016deep} & 55.1M & 89.01 & 77.48 & 51.64 & 72.07 & 74.43 & 77.55 & 87.76 & 90.81 & 79.71 & 85.27 & 58.36 & 64.11 & 76.50 & 71.99 & 54.06 &  74.05 \\
& \textbf{PKINet-S} & \textbf{44.8}M & 89.33 & 85.59 & 55.75 & 74.69 & 74.69 & 79.13 & 88.05 & 90.90 & 87.43 & 86.90 & 61.67 & 64.25 & 77.77 & 75.38 & 66.08 & \textbf{77.17}   \\
\hdashline
\multirow{3}{*}{\parbox{2cm}{\raggedleft Rotated Faster\\ R-CNN ~\cite{ren2015faster}}} & ResNet-50~\cite{he2016deep} & 41.1M & 89.40 & 81.81 & 47.28 & 67.44 & 73.96 & 73.12 & 85.03 & 90.90 & 85.15 & 84.90 & 56.60 & 64.77 & 64.70 & 70.28 & 62.22 & 73.17  \\
& ARC-R50~\cite{pu2023adaptive} & 74.4M & 89.49 & 82.11 & 51.02 & 70.38 & 79.07 & 75.06 & 86.18 & 90.91 & 84.23 & 86.41 & 56.10 & 69.42 & 65.87 & 71.90 & 63.47 & 74.77 \\
& \textbf{PKINet-S} & \textbf{30.8}M & 89.33 & 85.27 & 52.34 & 73.03 & 73.72 & 75.60 & 86.97 & 90.88 & 86.52 & 87.30 & 64.23 & 64.20 & 75.63 & 80.31 & 61.47 & \textbf{76.45} \\
\hdashline
\multirow{4}{*}{O-RCNN~\cite{xie2021oriented}} & ResNet-50~\cite{he2016deep} & 41.1M & 89.46 & 82.12 & 54.78 & 70.86 & 78.93 & 83.00 & 88.20 & 90.90 & 87.50 & 84.68 & 63.97 & 67.69 & 74.94 & 68.84 & 52.28 & 75.87 \\
& ARC-R50~\cite{pu2023adaptive} & 74.4M & 89.40 & 82.48 & 55.33 & 73.88 & 79.37 & 84.05 & 88.06 & 90.90 & 86.44 & 84.83 & 63.63 & 70.32 & 74.29 & 71.91 & 65.43 &  77.35 \\
& LSKNet-S~\cite{Li_2023_ICCV}  & 31.0M  & 89.66 & 85.52 & 57.72 & 75.70 & 74.95 & 78.69 & 88.24 & 90.88 & 86.79 & 86.38 & 66.92 & 63.77 & 77.77 & 74.47 & 64.82 & 77.49    \\
& \textbf{PKINet-S} & \textbf{30.8}M & 89.72 & 84.20 & 55.81 & 77.63 & 80.25 & 84.45 & 88.12 & 90.88 & 87.57 & 86.07 & 66.86 & 70.23 & 77.47 & 73.62 & 62.94 & \textbf{78.39} \\
\hline
\end{tabular}}
\vspace{-8pt}
\caption{\textbf{Experimental results on DOTA-v1.0 dataset} \cite{xia2018dota} under single-scale training and testing setting. PKINet-S is pretrained on ImageNet-1K \cite{deng2009imagenet} for 300 epochs similar to the compared methods \cite{xie2021oriented,ding2019learning,yang2021r3det}. See \S\ref{sec:quantitative_results} for details.}
\label{tab:var_arch}
\vspace{-6pt}
\end{table*}

\section{Experiment}
\subsection{Experimental Setup}
\noindent\textbf{Datasets.} We conduct extensive experiments on four popular remote sensing object detection datasets:
\begin{itemize}
    \item \textbf{DOTA-v1.0} \cite{xia2018dota} is a$_{\!}$ large-scale$_{\!}$ dataset$_{\!}$ for$_{\!}$ remote$_{\!}$ sensing$_{\!}$ detection$_{\!}$ which$_{\!}$ contains$_{\!}$ 2806$_{\!}$ images, 188,282 instances, and 15 categories with a large variety of orientations and scales.
    The dataset is comprised of 1,411, 458, and 937 images for \texttt{train}, \texttt{val}, and \texttt{test}, respectively.
    \item \textbf{DOTA-v1.5} \cite{xia2018dota} is a more challenging dataset based on DOTA-v1.0 which is released for DOAI Challenge 2019. This iteration includes the addition of a novel category named \texttt{Container Crane} (CC) and a substantial increase in the number of minuscule instances that are less than 10 pixels, containing 403,318 instances in total.
    \item \textbf{HRSC2016} \cite{liu2017high} is a remote sensing dataset for ship detection that contains 1061 aerial images whose size ranges from $300 \times 300$ and $1500 \times 900$. The images splits into 436/181/444 for \texttt{train/val/test.}
    \item \textbf{DIOR-R} \cite{cheng2022anchor} provides OBB annotations based on remote sensing dataset DIOR \cite{li2020object} dataset. It contains 23,463 images with the size of $800 \times 800$ and 192,518 annotations.
\end{itemize} 

\begin{table}[t]
\centering
\resizebox{0.49\textwidth}{!}{
			\setlength\tabcolsep{15pt}
			\renewcommand\arraystretch{1.2}
\hspace{-1.0em}
\begin{tabular}{c||c|c|c}
\spthickhline
\rowcolor[rgb]{0.92,0.92,0.92}Backbone  & \#\textbf{Params} $\downarrow$ & \#\textbf{FLOPs} $\downarrow$  & \textbf{mAP} $\uparrow$ \\ \hline \hline
ResNet-18 \cite{he2016deep} & 11.2M                                                 & 38.1G      & 74.20                                           \\
PKINet-T (\textbf{ours})  & \textbf{4.1}M                                                 & \textbf{22.7}G   & \textbf{77.87}                                             \\ \hline
ResNet-50 \cite{he2016deep} & 23.3M                                                 & 86.1G  & 75.87                                              \\
PKINet-S (\textbf{ours}) & \textbf{13.7}M                                                 & \textbf{70.2}G   & \textbf{78.39}                \\          
\hline
\end{tabular}
}
\vspace{-8pt}
\caption{\textbf{Comparison with ResNet} \cite{he2016deep} \textbf{backbone on DOTA-v1.0 dataset} \cite{xia2018dota}. Params and FLOPs are computed for backbones only. All the backbones are pretrained on ImageNet-1K \cite{deng2009imagenet} for 300 epochs and built within Oriented RCNN \cite{xie2021oriented}. See \S\ref{sec:quantitative_results} for details.}
\vspace{-14pt}
\label{tab:param_flops}
\end{table}

\begin{table*}[t]
\centering
\vspace{-5pt}
			\setlength\tabcolsep{5pt}
			\renewcommand\arraystretch{1.15}
\resizebox{\textwidth}{!}{
\begin{tabular}{r||cccccccccccccccc||c} 
\spthickhline
\rowcolor[rgb]{0.92,0.92,0.92} Method                            & PL    & BD    & BR    & GTF   & SV    & LV    & SH    & TC    & BC    & ST    & SBF   & RA    & HA    & SP    & HC  &CC & \textbf{mAP $\uparrow$}  \\ 
\hline
\hline
RetinaNet-O~\cite{lin2017focal} & 71.43 & 77.64 & 42.12 & 64.65 & 44.53 & 56.79 & 73.31 & 90.84 & 76.02 & 59.96 & 46.95 & 69.24 & 59.65 & 64.52 & 48.06 & 0.83 & 59.16 \\
FR-O~\cite{ren2015faster}  & 71.89 & 74.47 & 44.45 & 59.87 & 51.28 & 68.98 & 79.37 & 90.78 & 77.38 & 67.50 & 47.75 & 69.72 & 61.22 & 65.28 & 60.47 & 1.54 & 62.00 \\
Mask R-CNN~\cite{he2017mask}  & 76.84 & 73.51 & 49.90 & 57.80 & 51.31 & 71.34 & 79.75 & 90.46 & 74.21 & 66.07 & 46.21 & 70.61 & 63.07 & 64.46 & 57.81 & 9.42 & 62.67 \\
HTC~\cite{chen2019hybrid} & 77.80 & 73.67 & 51.40 & 63.99 & 51.54 & 73.31 & 80.31 & 90.48 & 75.12 & 67.34 & 48.51 & 70.63 & 64.84 & 64.48 & 55.87 & 5.15 & 63.40 \\
ReDet~\cite{han2021redet}  & 79.20 & 82.81 & 51.92 & 71.41 & 52.38 & 75.73 & 80.92 & 90.83 & 75.81 & 68.64 & 49.29 & 72.03 & 73.36 & 70.55 & 63.33 & 11.53 & 66.86 \\
DCFL~\cite{xu2023dynamic}  &- &- &- &- & 56.72 &- & 80.87 &- &- & 75.65 &- &- &- & - &- &- & 67.37 \\
LSKNet-S~\cite{Li_2023_ICCV}  & 72.05 & 84.94 & 55.41 & 74.93 & 52.42 & 77.45 & 81.17 &90.85 & 79.44 & 69.00 & 62.10 & 73.72 & 77.49 & 75.29 & 55.81 & 42.19 & 70.26 \\
\hline
\hline
PKINet-S (\textbf{ours}) & 80.31 & 85.00 & 55.61 & 74.38 & 52.41 & 76.85 & 88.38 & 90.87 & 79.04 & 68.78 & 67.47 & 72.45 & 76.24 & 74.53 & 64.07 & 37.13 & \textbf{71.47} \\
\hline
\end{tabular}
}
\vspace{-8pt}
\caption{\textbf{Experimental results on DOTA-v1.5 dataset} \cite{xia2018dota} compared with state-of-the-art methods with single-scale training and testing. PKINet-S backbone is pretrained on ImageNet-1K \cite{deng2009imagenet} for 300 epochs, as the compared methods \cite{han2021redet, xu2023dynamic}. PKINet-S is built within the framework of Oriented RCNN \cite{xie2021oriented}. See \S\ref{sec:quantitative_results} for details.}
\label{tab:main_performance_dotav15}
\vspace{-12pt}
\end{table*}

\noindent\textbf{Training.} Our training process contains ImageNet \cite{deng2009imagenet} pretrain and remote sensing object detector training. For ImageNet pretrain, our PKINet is trained on the ImageNet-1K under the MMPretrain \cite{2023mmpretrain} toolbox. In the main experiment, we train it for 300 epochs for higher performance like previous works \cite{xie2021oriented,yang2021r3det,pu2023adaptive, Li_2023_ICCV}. In the process of pretrain, we adapt the AdamW \cite{kingma2014adam} optimizer with a momentum of 0.9 and a weight decay of 0.05. Cosine schedule \cite{loshchilov2016sgdr} and warm-up strategy are employed to adjust the learning rate. We use 8 GPUs with a batch size of 1024 for pertaining. For remote sensing object detector training, experiments are conducted on MMRotate \cite{zhou2022mmrotate} framework. To compare with other methods, we use \texttt{trainval sets} of these benchmarks and their \texttt{test sets} for testing. Following the settings of previous methods \cite{han2021redet, xie2021oriented, yang2021r3det, zeng2023ars}, we crop original images into $1024 \times 1024$ patches with overlaps of 200 for DOTA-v1.0 and DOTA-v1.5 datasets. For HRSC2016 and DIOR-R datasets, the input size is set as $800 \times 800$. Models are trained with 30 epochs, 30 epochs, 60 epochs, and 36 epochs for DOTA-v1.0, DOTA-v1.5, HRSC2016, and DIOR-R. We employ AdamW \cite{kingma2014adam} optimizer with a weight decay of 0.05. The initial learning rate is set to 0.0002. All flops reported are calculated when the input image size is $1024 \times 1024$. To prevent over-fitting, images undergo random resizing and flipping during training following previous methods \cite{han2021redet, xie2021oriented, yang2021r3det, zeng2023ars}. Five-run average mAP of our method are reported for HRSC2016 and DIOR-R. 

\noindent\textbf{Testing.} The image resolution at the testing stage remains consistent with the training stage. For the sake of fairness, we do not apply any test-time data augmentation.

\noindent\textbf{Evaluation$_{\!}$ Metric.$_{\!}$} The$_{\!}$ mean$_{\!}$ average$_{\!}$ precision$_{\!}$ (mAP)$_{\!}$ and$_{\!}$ the Average Precision at 0.5 threshold ($\mathrm{AP}_{50}$) are reported.

\noindent\textbf{Reproducibility.$_{\!}$} Our$_{\!}$ algorithm$_{\!}$ is$_{\!}$ implemented$_{\!}$ in$_{\!}$ PyTorch. We use eight NVIDIA RTX 4090 GPUs for ImageNet pretraining and four NVIDIA Tesla V100 GPUs for downstream training and testing. 


\subsection{Quantitative Results}
\label{sec:quantitative_results}
\noindent\textbf{Performance on DOTA-v1.0} \cite{xia2018dota}. To begin with, we make a comparison with ResNet \cite{he2016deep} built within the framework
of Oriented RCNN \cite{xie2021oriented} on DOTA-v1.0 in Table \ref{tab:param_flops}. PKINet-T outperforms by \textbf{3.67\%} using only \textbf{36.7\%} of the parameters and \textbf{59.6\%} of the compute needed by ResNet-18. PKINet-S also excels, improving by \textbf{2.52\%} with just \textbf{58.8\%} of the parameters and \textbf{81.53\%} of the compute of ResNet-50. 

Our PKINet backbone, when paired with multiple detection architectures shown in Table \ref{tab:var_arch}, consistently outperforms ResNet-50 and more networks designed for 
the remote sensing detection task ($i.e.$, ARC \cite{pu2023adaptive} and LSKNet \cite{Li_2023_ICCV}). For one-stage architectures, our backbone is able to bring \textbf{2.41\%}/\textbf{6.19\%}/\textbf{3.71\%} mAP improvement compared with ResNet-50 for Rotated FCOS \cite{tian2019fcos}, R3Det \cite{yang2021r3det}, and S$^2$ANet~\cite{han2021align} respectively. 
Even when integrated with the classical S$^2$ANet~\cite{han2021align}, our method surpasses the previous approaches, achieving a performance of \textbf{77.83\%}. 
For two-stage architectures, PKINet also achieves remarkable gain (\textbf{3.12\%}/\textbf{3.28\%}/\textbf{2.23\%}). When equipped with the advanced detector Oriented RCNN \cite{xie2021oriented}, the performance reaches the superior \textbf{78.39\%} with a remarkable performance improvement for small categories compared with the previous best method LSKNet \cite{Li_2023_ICCV} (\textbf{5.3\%}/\textbf{5.76\%} for SV/LV). For the RA category that needs more contextual information, PKINet also achieves \textbf{6.46\%} enhancement compared with LSKNet.

\begin{table}[t]
\centering
			\setlength\tabcolsep{8pt}
			\renewcommand\arraystretch{1.10}
\resizebox{0.99\linewidth}{!}{
\begin{tabular}{r||c|c|c} 
\spthickhline
\rowcolor[rgb]{0.92,0.92,0.92} Method  & \#\textbf{Params} $\downarrow$ & \textbf{mAP (07)} $\uparrow$ & \textbf{mAP (12)} $\uparrow$ \\ 
\hline
\hline
DRN~\cite{pan2020dynamic}  &- &- & 92.70 \\
GWD~\cite{yang2021rethinking}  & 47.4M & 89.85 & 97.37 \\
Rol Trans.~\cite{ding2019learning}  & 55.1M & 86.20 &- \\
Gliding Vertex~\cite{xu2020gliding}   & 41.1M & 88.20 &- \\
CenterMap~\cite{long2021creating}  & 41.1M &- & 92.80 \\
AOPG~\cite{cheng2022anchor}  &- & 90.34 & 96.22 \\
R3Det~\cite{yang2021r3det} & 41.9M & 89.26 & 96.01 \\
S$^2$ANet~\cite{han2021align}  & 38.6M & 90.17 & 95.01 \\
ReDet~\cite{han2021redet}  & 31.6M & 90.46 & 97.63 \\
O-RCNN~\cite{xie2021oriented} & 41.1M & 90.50 & 97.60 \\
O-RepPoints~\cite{li2022oriented} & 36.6M & 90.38 & 97.26 \\
LSKNet~\cite{Li_2023_ICCV}  & 31.0M & 90.65 & 98.46 \\
\hline
\hline
PKINet-S (\textbf{ours}) & \textbf{30.8}M & \textbf{90.70} & \textbf{98.54} \\
\hline
\end{tabular}
}
\vspace{-8pt}
\caption{\textbf{Experimental results on HRSC2016 dataset} \cite{liu2017high}. PKINet-S is pretrained on ImageNet-1K  \cite{deng2009imagenet} for 300 epochs which is consistent with previous methods \cite{han2021redet, Li_2023_ICCV, lyu2022rtmdet} and built within the framework of Oriented RCNN \cite{xie2021oriented}. mAP (07/12): VOC 2007 \cite{voc2007}/2012 \cite{voc2012} metrics. See \S\ref{sec:quantitative_results} for details.}
\label{tab:performance_hrsc}
\vspace{-4pt}
\end{table}

\begin{table}[t]
    \centering
    			\setlength\tabcolsep{2pt}
			\renewcommand\arraystretch{1.15}
\hspace{-1.0em}
    \resizebox{0.99\linewidth}{!}{
    \begin{tabular}{c||cccc}  
    \spthickhline
    \cellcolor[gray]{0.92}Method & RetinaNet-O~\cite{lin2017focal} & FR-OBB~\cite{ren2015faster} & RT~\cite{ding2019learning} & LSKNet-S~\cite{Li_2023_ICCV}\\
    \hline
    \hline
    \cellcolor[gray]{0.92}\textbf{mAP} $\uparrow$  & 57.55 & 59.54  & 63.87 & 65.90\\
    \hline
        \hline
    \cellcolor[gray]{0.92}Method & GGHL~\cite{huang2022general} & Oriented Rep~\cite{xie2021oriented}  & DCFL~\cite{xu2023dynamic} & PKINet-S (\textbf{ours})\\
    \hline
    \hline
    \cellcolor[gray]{0.92}\textbf{mAP} $\uparrow$ & 66.48 & 66.71 & 66.80 & \textbf{67.03} \\
    \hline
    \end{tabular}}
    \vspace{-8pt}
    \caption{\textbf{Experimental results on DIOR-R dataset} \cite{li2020object}. Following previous methods \cite{xu2023dynamic, xie2021oriented, huang2022general}, PKINet-S is pretrained on ImageNet-1K \cite{deng2009imagenet} for 300 epochs and built within the framework of Oriented RCNN \cite{xie2021oriented}. See \S\ref{sec:quantitative_results} for details.}
    \vspace{-16pt}
    \label{tab:performance_diorr}
\end{table}

\begin{figure*}[t]
	\begin{center}
		\includegraphics[width = \linewidth]{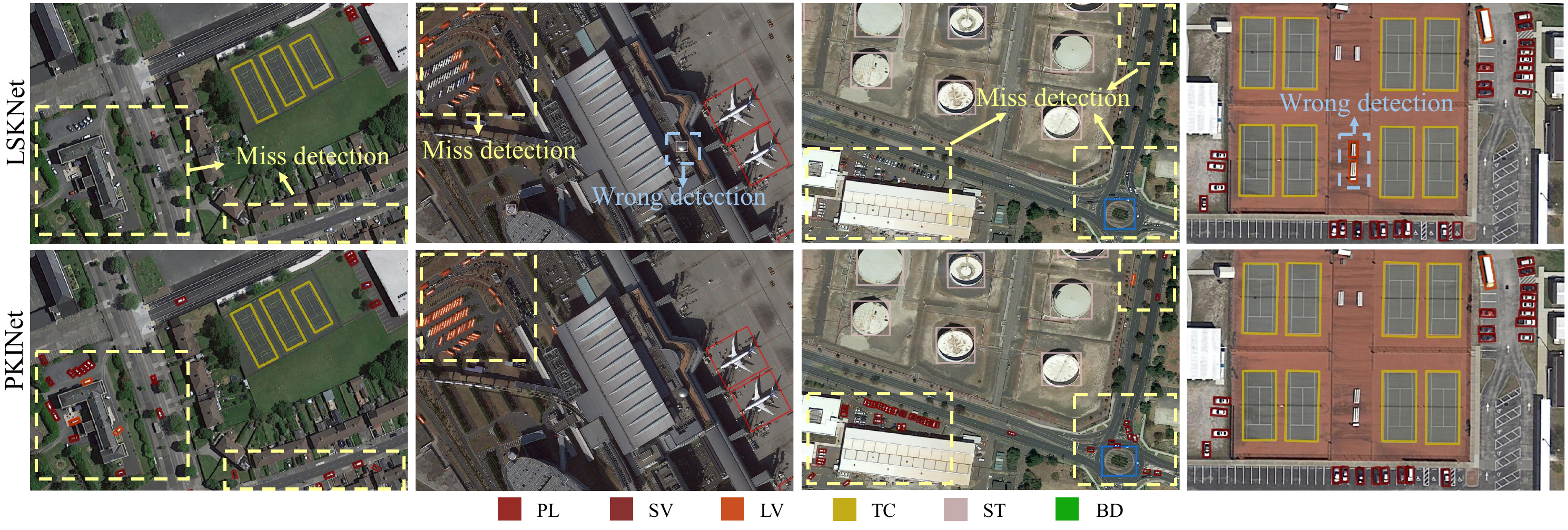}
	\end{center}
	\setlength{\abovecaptionskip}{0.cm}
        \vspace{-7pt}
	\caption{\textbf{Visual results on DOTA-v1.0 dataset} \cite{xia2018dota}. Top: LSKNet \cite{Li_2023_ICCV}; Bottom: our PKINet. See \S\ref{sec:qua_result} for details.}
	\label{fig:comparison}
 \vspace{-8pt}
\end{figure*}

\noindent\textbf{Performance on DOTA-v1.5} \cite{xia2018dota}. As shown in Table \ref{tab:main_performance_dotav15}, our approach achieves outstanding performance on the more challenging dataset DOTA-v1.5 with minuscule instances, evidencing its efficacy and generalization ability to small objects. Our PKINet outperforms the former state-of-the-art methods, achieving an improvement of \textbf{1.21\%}. 
C.

\noindent\textbf{Performance on HRSC2016} \cite{liu2017high}.
Our PKINet-S surpasses 12 leading methods on the HRSC2016 dataset with fewer parameters, as illustrated in Table \ref{tab:performance_hrsc}. The slight edge over LSKNet \cite{Li_2023_ICCV} mainly stems from HRSC2016 merging 31 subclasses into a single `ship' category for training and testing. This protocol doesn't fully showcase our method's strengths in managing inter-class object size variation.

\noindent\textbf{Performance on DIOR-R} \cite{cheng2022anchor}. We present comparison results on DIOR-R, as shown in Table \ref{tab:performance_diorr}. We achieve the best performance with \textbf{67.03\%}.

\noindent\textbf{Performance on COCO 2017} \cite{lin2014microsoft}. To assess the versatility of PKINet as a general framework adaptable to various forms of bounding boxes, we evaluate our method on the widely-used general detection benchmark COCO. As can be seen in Table 7, PKINet outperforms several famous backbones with similar parameters, thereby further afﬁrming the efﬁcacy of our method as a general-purpose backbone that is not confined to RSIs.

\vspace{-3pt}
\subsection{Qualitative Results}
\label{sec:qua_result}
Fig. \ref{fig:comparison} depicts representative visual results on DOTA \cite{xia2018dota}. As seen, compared to the previous best-performing method LSKNet \cite{Li_2023_ICCV} which merely relies on large kernels, our PKINet demonstrates a strong ability to adjust to significant size variations of target objects in a scene, ensuring detection of larger items ($e.g.$, PL, TC, ST, and BD) while retaining focus on smaller ones ($e.g.$, SV and LV).

\begin{table}[t]
    \centering
    			\setlength\tabcolsep{2pt}
			\renewcommand\arraystretch{1.15}
\hspace{-1.0em}
\resizebox{\linewidth}{!}{%
\setlength\tabcolsep{4pt}
\begin{tabular}{c||ccccc}
\sspthickhline
\cellcolor[gray]{0.92}Method   & ResNet-18~\cite{he2016deep} & PVT-T~\cite{wang2021pyramid} & ConvNeXt-N~\cite{liu2022convnet}  & PKINet-S (\textbf{ours})  \\ \hline \hline
\cellcolor[gray]{0.92}\#\textbf{Params} $\downarrow$ & \textbf{11.2}    & 12.9   & 15.6    & 13.7 \\
\hline
\cellcolor[gray]{0.92}\textbf{mAP} $\uparrow$ & 34.0    & 36.7   & 41.3   & \textbf{43.4} \\
\hline
\end{tabular}%
}
\vspace{-8pt}
    \caption{\textbf{Experimental results on COCO 2017 dataset} \cite{lin2014microsoft}. All models are pretrained on ImageNet-1K \cite{deng2009imagenet} for 300 epochs and are based on Mask R-CNN \cite{he2017mask}. See \S\ref{sec:quantitative_results} for details.}
\vspace{-12pt}
\label{tab:performance_coco}
\end{table}

\subsection{Diagnostic Experiments}
\label{sec:Dia_exp}
To gain more insights into PKINet, a set of ablative studies on DOTA-v1.0 is conducted with Oriented RCNN \cite{xie2021oriented} as the detector. All the backbones mentioned in this section are trained on ImageNet-1K \cite{deng2009imagenet} for 100 epochs for efficiency.

\noindent\textbf{Multi-scale Kernel Design.} First, the critical multi-scale kernel design in PKINet ($cf.$ \S\ref{sec:pki_module}) is investigated in Table \ref{tab:ms_kernel}. It demonstrates that using only small $3 \times 3$ kernels yields poor performance due to limited texture information extraction. Then a multi-scale kernel structure is adopted whose kernel size ranges from $3 \times 3$ to $11 \times 11$ with a stride of 2. Under this setting, the model shows the best performance. 
Next, a stride of 4 when the kernel size increases is tested and its performance is sub-optimal. Further trials with only large kernels led to increased computation but decreased performance, dropping by \textbf{0.49\%} and \textbf{0.84\%}, indicating that large kernels may introduce background noise and bring a performance drop ($cf.$ \S\ref{sec:intro}).

Then, we investigate the kernel number in multi-scale kernel design, detailed in \S\ref{sec:pki_module}. As Table \ref{tab:kernel_number} shows, with only two kernels (only $3\!\times\!3$ and $5\!\times\!5$ kernels are reserved), the network can't capture long-range pixel relationships. As the number of kernels rises, network performance improves, achieving optimal outcomes with five kernels.

\begin{table*}[t]
        \centering
	\begin{subtable}{0.35\linewidth}
		\resizebox{\textwidth}{!}{
			\setlength\tabcolsep{2pt}
			\renewcommand\arraystretch{1}
        \begin{tabular}{r||c|c|c}
        \spthickhline
        \rowcolor[rgb]{0.92,0.92,0.92} Kernel Design & \#\textbf{Params} $\downarrow$ & \#\textbf{FLOPs} $\downarrow$ & \textbf{mAP} $\uparrow$ \\ \hline \hline
        (3, 3, 3, 3, 3)  & \textbf{12.62}M & \textbf{62.40}G & 76.94 \\ \hline
        {\color{red}(3, 5, 7, 9, 11)} & 13.69M & 70.20G & \textbf{78.16} \\ \hline
        (3, 5, 9, 13, 17) & 14.99M & 79.57G & 78.07 \\ \hline
        (11, 11, 11, 11, 11) & 15.13M & 80.61G & 77.67 \\ \hline
        (15, 15, 15, 15, 15) & 17.44M & 92.45G & 77.32 \\ \hline
        \end{tabular}
		}
		\vspace{-9px}
		\setlength{\abovecaptionskip}{0.3cm}
		\setlength{\belowcaptionskip}{-0.1cm}
		\caption{multi-scale kernel design}
		\vspace{3px}
		\label{tab:ms_kernel}
	\end{subtable}
	\hspace{-0.7em}
        \begin{subtable}{0.33\linewidth}
		\resizebox{\textwidth}{!}{
			\setlength\tabcolsep{2pt}
			\renewcommand\arraystretch{1}
        \begin{tabular}{c||c|c|c}
        \spthickhline
        \rowcolor[rgb]{0.92,0.92,0.92} Kernel Number & \#\textbf{Params} $\downarrow$ & \#\textbf{FLOPs} $\downarrow$ & \textbf{mAP} $\uparrow$ \\ \hline \hline
        2  & \textbf{12.56}M & \textbf{61.95}G  & 75.76 \\ \hline
        3 & 12.78M & 63.57G & 76.07 \\ \hline
        4 & 13.13M & 66.24G & 77.53 \\ \hline
        {\color{red}5} & 13.69M & 70.20G & \textbf{78.16} \\ \hline
        6 & 14.35M & 75.26G & 78.10 \\ \hline
        \end{tabular}
		}
		\vspace{-9px}
		\setlength{\abovecaptionskip}{0.3cm}
		\setlength{\belowcaptionskip}{-0.1cm}
		\caption{kernel number}
		\vspace{3px}
		\label{tab:kernel_number}
	\end{subtable}
        \hspace{-0.7em}
        \begin{subtable}{0.31\linewidth}
		\resizebox{\textwidth}{!}{
			\setlength\tabcolsep{2pt}
			\renewcommand\arraystretch{1}
        \begin{tabular}{c||c|c|c}
        \sspthickhline
        \rowcolor[rgb]{0.92,0.92,0.92} Stage Apply& \#\textbf{Params} $\downarrow$ & \#\textbf{FLOPs} $\downarrow$ & \textbf{mAP} $\uparrow$ \\ \hline \hline
        None  & \textbf{12.03}M & \textbf{61.72}G  & 77.13 \\ \hline
        1 & 12.19M & 64.04G & 77.35 \\ \hline
        2 & 12.31M & 65.45G & 77.48 \\ \hline
        3 & 12.97M & 66.59G & 77.72 \\ \hline
        {\color{red}ALL} & 13.69M & 70.20G & \textbf{78.16} \\ \hline
        \end{tabular}
		}
		\vspace{-9px}
		\setlength{\abovecaptionskip}{0.3cm}
		\setlength{\belowcaptionskip}{-0.1cm}
		\caption{location for implementing CAA}
		\vspace{3px}
		\label{tab:caa_location}
	\end{subtable} \\
	\begin{subtable}{0.31\linewidth}
		\resizebox{\textwidth}{!}{
			\setlength\tabcolsep{10pt}
			\renewcommand\arraystretch{1.135}
        \begin{tabular}{c||c|c}
        \spthickhline
        \rowcolor[rgb]{0.92,0.92,0.92} Kernel Dilations & Max RF & \textbf{mAP} $\uparrow$ \\ \hline \hline
        {\color{red}(1, 1, 1, 1, 1)}  & 13  & \textbf{78.16} \\ \hline
        (2, 2, 2, 2, 2) & 24 & 77.07 \\ \hline
        (3, 3, 3, 3, 3) & 36 & 76.95 \\ \hline
        \end{tabular}
		}
		\vspace{-9px}
		\setlength{\abovecaptionskip}{0.3cm}
		\setlength{\belowcaptionskip}{-0.1cm}
		\caption{kernel dilations}
		\vspace{12px}
		\label{tab:dilation_kernel}
	\end{subtable}
         \hspace{-0.3em}
	\begin{subtable}{0.35\linewidth}
		\resizebox{\textwidth}{!}{
			\setlength\tabcolsep{2pt}
			\renewcommand\arraystretch{1.1}
        \begin{tabular}{c||r|c|c|c}
        \spthickhline
        \rowcolor[rgb]{0.92,0.92,0.92} CSP & \multicolumn{1}{c|}{Blocks} & \#\textbf{Params} $\downarrow$ & \#\textbf{FLOPs} $\downarrow$ & \textbf{mAP} $\uparrow$ \\ \hline \hline
        {\color{red}$\surd$} & (4, 12, 20, 4) & \textbf{13.69}M & 70.20G & \textbf{78.16} \\ \hline
        × & (4, 12, 20, 4) & 42.59M & 182.07G & - \\ \hline
        × & (2, 2, 4, 2) & 17.30M & \textbf{58.60}G & 77.83 \\ \hline
        \end{tabular}
		}
		\vspace{-9px}
		\setlength{\abovecaptionskip}{0.3cm}
		\setlength{\belowcaptionskip}{-0.1cm}
		\caption{cross-stage partial structure}
		\vspace{12px}
		\label{tab:csp}
	\end{subtable}
        \hspace{-0.3em}
	\begin{subtable}{0.32\linewidth}
		\resizebox{\textwidth}{!}{
			\setlength\tabcolsep{2pt}
			\renewcommand\arraystretch{1.1}
        \begin{tabular}{r||c|c|c}
        \spthickhline
        \rowcolor[rgb]{0.92,0.92,0.92} Kernel Design & \#\textbf{Params} $\downarrow$ & \#\textbf{FLOPs} $\downarrow$ & \textbf{mAP} $\uparrow$ \\ \hline \hline
        (3, 3, 3)  & \textbf{13.50}M & \textbf{68.95}G & 77.52 \\ \hline
        (5, 5, 5) & 13.52M & 69.08G & 77.71 \\ \hline
        (5, 7, 7) & 13.54M & 69.21G & 77.76 \\ \hline
        (7, 11, 11) & 13.58M & 69.47G & 77.89 \\ \hline
        {\color{red}Expansive}  & 13.69M & 70.20G & \textbf{78.16} \\ \hline
        \end{tabular}
		}
		\vspace{-9px}
		\setlength{\abovecaptionskip}{0.3cm}
		\setlength{\belowcaptionskip}{-0.1cm}
		\caption{kernel size in CAA}
		\vspace{12px}
		\label{tab:caa_kernel}
	\end{subtable}
 \vspace{-16pt}
\caption{\textbf{A set of ablative studies on DOTA-v1.0} \cite{xia2018dota}. The adopted network designs are marked in {\color{red}red}. All the networks are pretrained on ImageNet-1K \cite{deng2009imagenet} for 100 epochs and built with the framework of Oriented RCNN \cite{xie2021oriented}. See \S \ref{sec:Dia_exp} for details.}
\vspace{-8pt}
\end{table*}

\noindent\textbf{Kernel Dilations.} Then, we examine the effect of dilations in our PKI module ($cf.$ \S\ref{sec:pki_module}). 
As displayed in Table \ref{tab:dilation_kernel}, there is a performance degradation (-\textbf{1.09\%}) despite the increase in the receptive field compared to no kernel dilations. As we further increase the degree of dilation, a further drop in performance occurs. This proves that merely applying dilation to expand the receptive field does not work.

\noindent\textbf{Context Anchor Attention.} Next, the effectiveness of CAA module ($cf.$ \S\ref{sec:caa}) is proved. To start with, CAA is applied with different kernel sizes to check the impact in Table \ref{tab:caa_kernel}. The three kernel sizes in the first column represent the size in average pooling and two strip convolutions. As can be seen, smaller kernels fail to capture long-range dependencies, reducing performance, while larger kernels improve this by including more context. Our expansive kernel size strategy that increases the kernel size of strip convolutions as the blocks deepen achieves the best performance.

After that, since there are four stages in our PKINet, how the implementing location affects the final performance is investigated. As revealed in Table \ref{tab:caa_location}, CAA module ($cf.$ \S\ref{sec:caa}) can bring performance improvement when implemented at any stage. Consequently, when deploying CAA module at all stages, the performance gain reaches \textbf{1.03\%}.

\noindent\textbf{Cross-Stage Partial Structure.} Table \ref{tab:csp} further explores the impact of the Cross-Stage Partial (CSP) structure. Eliminating CSP leads to exponential increases in both parameters and computational costs (by \textbf{211\%} and \textbf{159\%}, respectively). By reducing the number of blocks in each stage from (4, 12, 20, 4) to (2, 2, 4, 2), models without CSP structure can reach a similar parameter as the former one. However, sub-optimal performance is achieved due to the decreased number of stage blocks.

\subsection{Analysis}
\label{sec:analysis}
To measure the model's detection sensitivity concerning the sizes of different categories, we utilize Pearson Correlation Coefficient (PCC)~\cite{cohen2009pearson} to quantify the linear correlation between the average bounding box area per category and the average detection score per category of DOTA-v1.0 \cite{xia2018dota}.

First, we calculate the average area of all the annotations for the $k$-th category, donated as $S_k$. The average area for all categories $\Bar{S}$ is calculated as $\bar{S}\!=\!\frac{1}{K} \sum_{k\!=\!1}^K S_k$, where $K$ is the number of categories. The mean scores for each category $Q_k$ and for all categories $\Bar{Q}$ are computed in a similar manner. Second, we calculate the covariance between the category-wise average areas $\{S_k\}_{k\!=\!1}^K$ and the category-wise average scores $\{Q_k\}_{k\!=\!1}^K$ as $D\!=\!\frac{1}{(K\!-\!1)}\sum\nolimits_{k\!=\!1}^K(S_k\!-\!\bar{S})\!\times\!(Q_k\!-\!\bar{Q})$. 
Finally, PCC is computed as:
\vspace{-4pt}
\begin{equation}\small
r = {D}/{\sigma_S \sigma_Q}.
\vspace{-4pt}
\end{equation}
Here, $\sigma_S$ and $\sigma_Q$ are the standard deviations of the category-wise average areas $\{S_k\}_{k=1}^K$ and the category-wise average scores $\{Q_k\}_{k=1}^K$, respectively. 
A PCC absolute value $|r|$ close to $0$ suggests a minimal linear correlation, indicating that the model's detection performance is rarely influenced by the size of the object. 
As illustrated in Table~\ref{tab:PCC}, our PKINet achieves both the highest mAP and the lowest PCC absolute value $|r|$, indicating that PKINet is the least sensitive to size variations across different categories.

\begin{table}[t]
\centering
			\setlength\tabcolsep{22pt}
			\renewcommand\arraystretch{1.2}
   \resizebox{0.4\textwidth}{!}{
\hspace{-1.0em}
\begin{tabular}{c||c|c}
\spthickhline
\rowcolor[rgb]{0.92,0.92,0.92}Methods & {mAP $\uparrow$}   & \textbf{ $|r|~\downarrow$}  \\ \hline \hline
S$^2$ANet~\cite{han2021align}  & 74.13 & 0.23 \\ \hline
O-RCNN~\cite{xie2021oriented}  & 75.87 & 0.22 \\ \hline
ARC~\cite{hou2022shape}     & 77.35 & 0.29 \\ \hline
LSKNet~\cite{Li_2023_ICCV}  & 77.49 & 0.24 \\ \hline
PKINet-S (\textbf{ours})  & \textbf{78.39} & \textbf{0.19} \\
\hline
\end{tabular}
}
\vspace{-8pt}
\caption{\textbf{Comparison of mAP and PCC ($r$) on DOTA-v1.0} dataset \cite{xia2018dota}. See \S\ref{sec:analysis} for details.}
\vspace{-12pt}
\label{tab:PCC}
\end{table}


\section{Discussion and Conclusion}
In this paper, we propose Poly Kernel Inception Network (PKINet) for remote sensing object detection, which aims at tackling the challenges posed by considerable variations in object scale and contextual diversity in remote sensing images. PKINet employs parallel depth-wise convolution kernels of various sizes to capture dense texture features effectively across different scales. A Context Anchor Attention mechanism is also introduced to capture long-range contextual information further. We experimentally show that PKINet achieves state-of-the-art performance on four famous remote sensing benchmark datasets.

\vspace{3mm}
\noindent\textbf{Limitations and Future Work.} While both PKINet-T and PKINet-S have demonstrated superior detection performance over previous methods, limitations in our computational resources have restricted PKINet from scaling up the model capacity to achieve its maximal potential. Similar studies on model scalability have received substantial interest in general object detection, as highlighted in Swin Transformer~\cite{liu2021swin} and ConvNeXt~\cite{liu2022convnet}. We leave further investigation into the scalability of PKINet for future research.

\clearpage
{
    \small
    \bibliographystyle{ieeenat_fullname}
    \bibliography{main}
}


\end{document}